\documentclass[10pt, a4paper]{article}

\usepackage{lrec-coling2024} 
\usepackage{latexsym}
\usepackage{textcomp}
\usepackage{booktabs}
\usepackage{subcaption}
\usepackage{tikz-dependency}
\usepackage{graphicx}
\graphicspath{ {./prodigy/} }
\usepackage{graphicx}
\usepackage{makecell}
\usepackage{tabularx}
\usepackage{multirow}
\usepackage{adjustbox}
\usepackage{mathtools}
\usepackage{enumitem}
\usepackage{algorithm2e}
\usepackage{csquotes}
\usepackage{courier}
\usepackage{tcolorbox}
\usepackage{soul}
\usepackage{setspace}
\usepackage{hyperref}
\usepackage{fancyvrb}
\title{Linear Cross-document Event Coreference Resolution with X-AMR}

\name{Shafiuddin Rehan Ahmed,\; George Arthur Baker,\; Evi Judge,\; Michael Regan$^\dagger$ \\[1mm] {\large \textbf{Kristin Wright-Bettner,\; Martha Palmer,} and \textbf{James H. Martin}}\\[-3mm]}

\address{University of Colorado, Boulder, CO, USA \hspace{5mm} $^\dagger$University of Washington, Seattle, WA, USA \\ \texttt{\{shah7567, george.baker\}@colorado.edu}}

\abstract{
Event Coreference Resolution (ECR) as a pairwise mention classification task is expensive both for automated systems and manual annotations. The task's quadratic difficulty is exacerbated when using Large Language Models (LLMs), making prompt engineering for ECR prohibitively costly. In this work, we propose a graphical representation of events, X-AMR, anchored around individual mentions using a \textbf{cross}-document version of \textbf{A}bstract \textbf{M}eaning \textbf{R}epresentation. We then linearize the ECR with a novel multi-hop coreference algorithm over the event graphs. The event graphs simplify ECR, making it a) LLM cost-effective, b) compositional and interpretable, and c) easily annotated. For a fair assessment, we first enrich an existing ECR benchmark dataset with these event graphs using an annotator-friendly tool we introduce. Then, we employ GPT-4, the newest LLM by OpenAI, for these annotations. Finally, using the ECR algorithm, we assess GPT-4 against humans and analyze its limitations. Through this research, we aim to advance the state-of-the-art for efficient ECR and shed light on the potential shortcomings of current LLMs at this task. Code and annotations: \href{https://github.com/ahmeshaf/gpt_coref}{\tt{https://github.com/ahmeshaf/gpt\_coref}}
 \\ \newline  }

\begin{document}

\maketitleabstract

\newcommand{\prop}{\text{PropBank}}
\newcommand{\mone}{\ensuremath{m_{1}}}
\newcommand{\mtwo}{\ensuremath{m_{2}}}
\newcommand{\mthree}{\ensuremath{m_{3}}}
\newcommand{\mfour}{\ensuremath{m_{4}}}
\newcommand{\mfive}{\ensuremath{m_{5}}}

\newcommand{\eone}{\ensuremath{e_{1}}}
\newcommand{\etwo}{\ensuremath{e_{2}}}
\newcommand{\ethree}{\ensuremath{e_{3}}}

\newcommand{\ecb}{\text{ECB+}}

\newcommand{\ecbone}{\textbf{39\_11ecbplus}}
\newcommand{\ecbtwo}{\textbf{39\_1ecb}}
\newcommand{\ecbthree}{\textbf{39\_5ecbplus}}
\newcommand{\replace}{\textcolor{blue}{$replace_{\mone}$}}
\newcommand{\replacing}{\textcolor{magenta}{$replacing_{\mtwo}$}}

\newcommand{\takesover}{\textcolor{blue}{$takes~over_{\mthree}$}}

\newcommand{\steppedinto}{\textcolor{blue}{$stepped~into_{\mfour}$}}

\newcommand{\blue}[1]{\textcolor{blue}{\textit{#1}}}
\newcommand{\evtone}{\blue{$e_{1}$}}
\newcommand{\evttwo}{\textcolor{magenta}{$e_{2}$}}
\newcommand{\evtthree}{\blue{$e_{3}$}}
\newcommand{\evtfour}{\blue{$e_{4}$}}

\tcbset{before upper={\linespread{0.1}}}
\definecolor{hlcolor}{HTML}{DAE6FB}
\definecolor{evtcolor}{HTML}{486485}
\sethlcolor{hlcolor}
\definecolor{mygray}{HTML}{808080}
\newtcolorbox{mybox1}[2][]{boxsep=1mm, arc=.15em, left=5pt,right=5pt,top=2pt,bottom=2pt,boxrule=0.1mm,
colbacktitle=blue!20!white,colback=white, coltitle=black,
title={#2}, #1}

\definecolor{arg0color}{HTML}{7FFFD4}
\definecolor{arg1color}{HTML}{98FB98}
\definecolor{arg2color}{HTML}{dda0dd}
\definecolor{argtcolor}{HTML}{87CEFA}
\definecolor{arglcolor}{HTML}{DCDCDC}
\definecolor{rscolor}{HTML}{ACACAC}

\newtcolorbox{mybox2}[2][]{boxsep=1mm, arc=.15em, left=5pt,right=5pt,top=2pt,bottom=2pt,boxrule=0.1mm,
colbacktitle=blue!45!white,colback=white, coltitle=white,
title={#2}, #1}

\newtcolorbox{mybox3}[2][]{boxsep=1mm, arc=.15em, left=5pt,right=5pt,top=2pt,bottom=2pt,boxrule=0.1mm,
colbacktitle=black!45!white,colback=white, coltitle=white,
title={#2}, #1}
\newcommand{\highlightevt}[1]{\hl{~\textbf{#1} {\textbf{\tiny {\color{evtcolor} EVT}}}~}}
\newcommand{\helvetext}[1]{\fontfamily{phv}\fontsize{7.5pt}{8pt}\selectfont  #1}
\newcommand{\helvetexttitle}[1]{\fontfamily{phv}\fontsize{8pt}{8pt}\selectfont \textbf{#1}}
\newcommand{\todo}[1]{\textcolor{red}{\hl{TODO: #1}}}
\newcommand{\argzero}{\ensuremath{\text{ARG-0}}}
\newcommand{\argone}{\ensuremath{\text{ARG-1}}}
\newcommand{\argL}{\ensuremath{\text{ARG-Loc}}}
\newcommand{\argT}{\ensuremath{\text{ARG-Time}}}

\section{Introduction}
Event Coreference Resolution (ECR) involves ide\-ntifying events that refer to the same real-world occurrence both within and across documents. Traditionally, ECR is performed on pairs of event mentions in a corpus through the use of rules, features, or neural methods to generate similarity scores \cite{kenyon2018resolving}, with neural methods such as Transformer-based encoders \cite{devlin-etal-2019-bert, liu2019roberta, beltagy2020longformer} achieving state-of-the-art performance on various ECR benchmarks \cite{caciularu-etal-2021-cdlm-cross, held-etal-2021-focus}. However, the quadratic nature of pairwise approaches makes it challenging to scale up to large corpora of thousands of documents.

Figure \ref{fig:examples} presents three event mentions (\mone, \mtwo, and \mthree) with their respective event triggers highlighted. \mone~and \mtwo~are examples of coreferent events, while \mthree~is a related yet non-coreferent event. While \mone~and \mthree~contain sufficient information required to make a negative coreferencing decision between them, additional extrasentential context is needed to determine the coreferential relationship between \mtwo~and the other two mentions. 
\begin{figure}[t]
    \centering
    \begin{mybox1}{\helvetexttitle{Target Mention (\mone)}}
{ \helvetext{HP today announced that it has signed a definitive agreement to \highlightevt{acquire} EYP Mission Critical Facilities Inc.}}
\end{mybox1}
\vspace*{-3.mm}
\begin{mybox1}{\helvetexttitle{Target Mention (\mtwo)}}
{\helvetext{Financial details of the \highlightevt{acquisition} were not disclosed.}}
\end{mybox1}
\vspace*{-3.mm}
\begin{mybox2}{\helvetexttitle{Target Mention (\mthree)}}
{\helvetext{Earlier this month Hewlett-Packard unveiled a bid of nearly \$14 billion bid to \highlightevt{purchase} Electronic Data Systems.}}
\end{mybox2}
    \vspace*{-3.0mm}
    \caption{\mone~and \mtwo~are examples of coreferent mentions. \mthree~although related to \mone~and \mtwo, is a different acquisition event.}
    \label{fig:examples}
    \vspace*{-5mm}
\end{figure}

\vspace{-4mm}

The challenge of ECR stems from the inherent issue of establishing \textit{singular terms} for event mentions ($a$, $b$) that can be compared for identity  (\textit{is a = b?}; \citet{Davidson1969}). Consequently, pairwise methods resort to approximations of the coreference relationship between each mention pair by leveraging either the sentence or the entire document for contextual information. The methods that need to rely on the entire document for each pairwise decision (as in the case of \mtwo) are intractable on large corpora. We propose that by extracting the key semantics of mentions and by introducing a graphical structure between each mention, we can compress the information. This way we not only are able to create identifiers for the mentions that can be compared for sameness, but also make ECR completely linear in complexity. 

\begin{figure}[b]
\vspace*{-4mm}
    \centering
    \resizebox{1\linewidth}{!}{\includegraphics{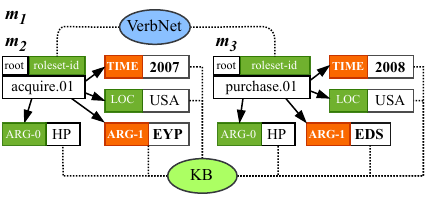}}
    \vspace*{-5mm}
    \caption{Compressed  event semantics graphs of \mone, \mtwo, and \mthree. The graph for \mtwo~is generated using the entire document in which it appeared. VerbNet classes are used for synonymous predicates. KB is used for argument coreference resolution.}
    \label{fig:coref_eg_standard}
    \vspace*{-3mm}
\end{figure}

Figure \ref{fig:coref_eg_standard} illustrates the graph structure based on our new Cross-document Abstract Meaning Representations (X-AMR), inspired by Abstract Meaning Representation (AMR; \citet{banarescu-etal-2013-abstract}).  X-AMR captures event triggers and arguments linked using  VerbNet Lexicon \citep{schuler2005verbnet} and a Knowledge Base (KB), resulting in corpus-level event graphs that we use to symbolically perform ECR without the need for pairwise scoring. Our specific contributions in this paper include:

\begin{itemize}[itemsep=0.3pt]
    \item X-AMR annotations using the annotation tool provided by \citet{ahmed-etal-2024-xtool} on the ECB+ corpus \cite{cybulska-vossen-2014-using}.
    \item A novel ECR algorithm over the X-AMR graphs that avoids the computational cost of traditional pairwise approaches. 
    \item An evaluation of the algorithm using gold-standard annotations for the ECB+ corpus. 
    \item And finally, an evaluation of the approach with automatically generated X-AMR graphs using GPT-4 in zero-shot and few-shot settings with prompts based on a condensed version of the annotation guidelines of X-AMR. 
\end{itemize}
Together, our annotations and findings suggest a promising path toward extending robust ECR to real-world applications where exhaustive pairwise approaches are not feasible. 
\vspace*{-1.3mm}
\section{Annotation Guidelines for X-AMR}
\vspace*{-1.2mm}
\label{sec:annotation}

We aim to annotate key event semantics with four arguments, \argzero, \argone, \argL, and \argT, capturing agent, patient (and theme), location, and temporal information. The selection of these arguments is to circumscribe an event by its \textit{minimal participants} \cite{lombard2019events, Guarino2022-GUAETN}.
We use the guidelines presented in the next section to hand annotate the roleset and argument information for the ECB+ train, development, and test sets using the standardized split of \citet{cybulska-vossen-2014-using}. Following the annotation guidelines, we provide the enriched annotations of the ECB+ corpus by two Linguistic students. We use the prodi.gy-based X-AMR annotation tool provided by \citet{ahmed-etal-2024-xtool}\footnote{Readers are encouraged to check the original paper for details about the annotation tool}. 
\vspace{-0.7mm}
\subsection{PropBank \& AMR}
\vspace{-0.5mm}
Semantic role labeling (SRL) centers on the task of assigning the same semantic role to an argument across various syntactic constructions. For example, \textit{the window} can be the (prototypical) Patient, or thing broken, whether expressed as syntactic object (\textit{The storm broke the window}) or syntactic subject (\textit{The window broke in the storm}). 

   
The Proposition Bank (PropBank; \citet{palmer-etal-2005-proposition,pradhan-etal-2022-propbank}) has over 11,000 Frame Files providing valency information (arguments and their descriptions) for fine-grained senses of English verbs, eventive nouns, and adjectives. Figure \ref{fig:enter-label} gives an example Frame File for \textit{agree} as well as an instantiated frame for \textit{HP has an agreement to acquire EYP}.  



\vspace{1.5mm}


\begin{figure}
    \centering
\begin{minipage}[b]{0.50\columnwidth}
\boxsize
\noindent
\begin{Verbatim}[baselinestretch=1.5]
agree.01 - agree
   ARG-0: Agreer
   ARG-1: Proposition
   ARG-2: Other entity 
              agreeing
\end{Verbatim}
\end{minipage}
\begin{minipage}[b]{0.01\columnwidth}
\hfill
\end{minipage}
\begin{minipage}[b]{0.43\columnwidth}
\boxsize
\begin{Verbatim}[baselinestretch=1.5]
agree.01
   ARG-0: HP 
   ARG-1: acquire.01 
      ARG-0: HP        
      ARG-1: EYP
\end{Verbatim}
\end{minipage}

    \caption{The PropBank roleset definitions of agree.01 and the expected annotations in X-AMR.}
    \label{fig:enter-label}
\vspace*{-2mm}
\end{figure}

The resulting nested predicate-argument structures from PropBank style-SRL also form the backbones of AMRs, which in addition includes Named Entity (NE) tags and Wikipedia links (for `HP' and `EYP' in our example). AMRs also include explicit variables for each entity and event, consistent with Neo-Davidsonian event semantics, as well as inter- and intra-sentential coreference links to form directed, (largely) acyclic graphs that represent the meaning of an utterance or set of utterances.
  
Our enhanced X-AMR representation follows AMR closely with respect to NE and coreference, but stops short of AMR's additional structuring of noun phrase modifiers (especially with respect to dates, quantities and organizational relations), the discourse connectives and the partial treatment of negation and modality. However, we go further than AMR by allowing for cross-document coreference as well as multi-sentence coreference.  X-AMR thus provides us with a flexible and expressive event representation with much broader coverage than standard event annotation datasets such as ACE\footnote{https://www.ldc.upenn.edu/collaborations/past-projects/ace} or Maven \cite{wang-etal-2020-maven}.
\vspace{-0.5mm}
\subsection{Roleset Sense Annotation}
The first step in the annotation process involves identifying the roleset sense for the target event trigger in the given text. Annotators, using an embedded PropBank website and the assistance of the tool's model, select the most appropriate sense by comparing senses across frame files. 

\noindent \textbf{Handling Triggers with No Suitable Roleset:}
If there is no appropriate roleset that specifies the event trigger, particularly in cases when the trigger is a pronoun (it) or proper noun (e.g., Academy Awards), the annotator must then search for a roleset that defines the appropriate predicate.


\subsection{Document-level Arguments Identification}
\vspace{-1mm}
Next, we identify the document and corpus-level \argzero~and \argone~of the selected roleset. Annotators use the embedded PropBank website as a reference for the roleset's definition, ensuring that the \argzero~(usually the agent) and \argone~(typically the patient) are consistent with the roleset's constraints. For arguments that cannot be inferred, the annotators leave those fields empty.

\begin{figure}[t]
\centering
\begin{mybox3}{\helvetexttitle{Target Mention (\mfour)}}
{ \helvetext{HP today announced that it has \highlightevt{signed} a definitive agreement to acquire EYP Mission Critical Facilities Inc.}}
\end{mybox3}
\includegraphics[scale=0.80]{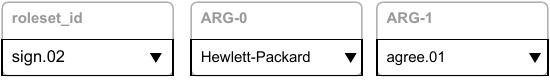}
\caption{Eventive \argone\space in \mfour~for the roleset sign.02. The \argone~clause is annotated as the connecting event with roleset ID agree.01}
\vspace*{-3.5mm}
\label{fig:eventive-arg}
\end{figure}

\vspace{1.3mm}
\noindent \textbf{Within- and Cross-Document Entity Coreference Annotation:}
Annotators perform within- and cross-document entity coreference using a drop-down box of argument suggestions (suggested by the model-in-the-loop), simplifying coreference link establishment. In difficult cases like \mtwo~(Fig \ref{fig:examples}), where \argzero\space and \argone\space are missing, the drop-down box helps by suggesting ``HP" and ``EYP" from the \mone~sentence. Similarly, in \mfour~(Figure \ref{fig:eventive-arg}), the drop-down box assists in resolving \argzero\space (it) as ``HP", using the information earlier within the sentence. Annotators are also allowed to input multiple values separated by ``/" as needed, (e.g.,  if two people performed some action together, "Person 1/Person 2").

\vspace{1.2mm}
\noindent \textbf{Nested \argone:}
In many cases, the \argone~may itself be an event. In such cases, the annotator is tasked with identifying the head predicate of the \argone~role and providing its corresponding roleset ID. We then search for the annotations for such an \argone~and connect it to the target event. Fig \ref{fig:eventive-arg} has an example of a mention with an eventive \argone. For this, the annotator needs to provide the roleset for the  predicate of the \argone~clause (agree.01) as the \argone\space in this annotation process.
\vspace{-3.5mm}
\paragraph{\argL~\& \argT~Identification} Annotators may also utilize external resources, such as Wikipedia\footnote{Although we add this in the guidelines, the annotators do not wikify. This is only for GPT to generate instructions for itself. Our choice is to use Wikipedia over the more commonly used KB-wikidata because of GPT-friendly identifiers of the pages.}, or Google-News, for
the accurate identification of temporal and spatial arguments. This is required when the document does not explicitly mention the location and time of the event.

\newcommand{\HP}{\ensuremath{\mathrm{HP}}}
\newcommand{\EYP}{\ensuremath{\mathrm{EYP}}}
\newcommand{\pbSyns}{\ensuremath{\text{}_{\text{syn}}}}
\newcommand{\PBSynHum}{\ensuremath{\mathtt{HUM}}}
\newcommand{\PBSynVN}{\ensuremath{\mathtt{VN}}}
\newcommand{\RS}{\ensuremath{\mathtt{PB}}}
\newcommand{\PB}{\ensuremath{\mathtt{PB}}}
\newcommand{\RSHum}{\ensuremath{\mathtt{PB_{H}}}}
\newcommand{\RSGPT}{\ensuremath{\PB_{\mathtt{G}}}}
\newcommand{\eid}{\ensuremath{\texttt{EID}}}
\newcommand{\eidnHop}{\ensuremath{\eid_{n}}}
\newcommand{\eidnminusoneHop}{\ensuremath{\eid_{n - 1}}}
\newcommand{\eidNHop}{\ensuremath{\eid_{\text{N}}}}
\newcommand{\eidNH}{\ensuremath{\eidNHop^{\mathtt{}}}}
\newcommand{\eidHVN}{\ensuremath{\eidNHop^{\mathtt{VN}}}}
\newcommand{\eidLEM}{\ensuremath{\eidNHop^{\mathtt{LEM}}}}
\newcommand{\eidLT}{\ensuremath{\eid_{\mathrm{lt}}}}
\newcommand{\eidLTH}{\ensuremath{\eidLT^{\mathtt{}}}}
\newcommand{\eidLTHVN}{\ensuremath{\eidLT^{\mathtt{VN}}}}
\newcommand{\eidLTLEM}{\ensuremath{\eidLT^{\mathtt{LEM}}}}
\newcommand{\eidtwohop}{\ensuremath{\eid_{2}}}
\newcommand{\eidonehop}{\ensuremath{\eid_{1}}}
\newcommand{\eidzerohop}{\ensuremath{\eid_{0}}}
\newcommand{\eidhop}{\ensuremath{\texttt{EID}_{k, n}^{\text{hop}}}}
\newcommand{\argRes}[1]{\ensuremath{\mathtt{Resolver}_{\text{arg}}}(#1)}
\newcommand{\pbRes}[1]{\ensuremath{\mathtt{Resolver}_{\text{pb}}}(#1)}
\newcommand{\concat}{\ensuremath{,}}
\vspace{-2.5mm}
\section{Human Annotations}
\vspace{-1.5mm}
To perform the X-AMR annotations, we employ two annotators, and we execute this process in a systematic two-step approach. In the initial phase, these annotators are responsible for identifying the roleset ID associated with each event trigger. We aggregate all event mentions for which both annotators have concurred on the same roleset ID. For those instances where there is a lack of consensus between the annotators, we enlist the assistance of an adjudicator to resolve the discrepancies.  The annotations that have been finalized, either through agreement or adjudication, are then collectively advanced to the subsequent task of identifying the arguments.
\vspace{-2mm}
\subsection{Annotation Analysis}
We have currently annotated all the mentions in the corpus with their Roleset IDs and 5,287 out of the 6,833 with X-AMR. In the three splits, only the Dev set has been fully annotated. We calculate the inter-annotator agreement (IAA) on the common Roleset predictions. The IAA is highest for the Dev set at 0.91, as depicted in Table \ref{tab:ecb}. Consequently, we utilize the Dev set as our benchmark for experiments in the following sections.

\renewcommand{\arraystretch}{1.4}
\begin{table}[htb]
\centering
\small
    \begin{tabular}{@{}c|ccccc|c@{}}  
     \toprule 
    \multicolumn{3}{c}{~} & \textbf{Train} & \textbf{Dev} & \multicolumn{1}{c}{\textbf{Test}} &\\ 

    \cline{1-7}
	&Documents && 594 & 196 & 206   &\\ 
	&Mentions  && 3808 &  1245 & 1780& \\ 
        \cline{2-6}
        &\makecell{ \\[-3mm] Roleset ID \\ Agreement} && 0.84 & \textbf{0.91} & 0.80 &\\
        \cline{2-6}
        &w/ X-AMR && 3195$^{*}$ & 1245 & 847$^{*}$ &\\ 
        &w/ Nested ARG-1 && 1081 & 325 & 220 &\\ 
        &w/ ARG-Loc && 2949 & 1243 & 707 &\\ 
        &w/ ARG-Time && 3192 & 1244 & 805 &\\ \cline{1-7}
        \bottomrule
	\end{tabular}
  \caption[\ecb~Corpus Statistics]{
	Corpus statistics for event mentions in \ecb~and the mentions annotated with X-AMR ($^*$Annotation in Progress). Inter-annotator agreement for the Roleset ID is highest for the Dev set. }
\label{tab:ecb}
\end{table}

\vspace{-1mm}
\noindent \textbf{Arguments}:
Our analysis reveals a significant presence of mentions with nested ARG-1 annotations, as highlighted in Table \ref{tab:ecb} (w/ Nested ARG-1). This underscores the importance of capturing nested event relationships effectively. Additionally, our annotations for location and time modifiers successfully capture this information for nearly all mentions (w/ X-AMR), thanks to the assistance provided by drop-down options and the model-in-the-loop approach. These tools are particularly valuable in cases where date references are not explicitly mentioned in the document.

\section{Graph-based ECR Algorithm}
Our proposed approach for ECR builds upon previous research efforts that use minimum participants. \citet{cybulska-vossen-2013-semantic} utilize heuristics to ascertain event relationships based on various factors, such as location, time, and participant compatibility. \citet{choubey-huang-2017-event} employ iterative techniques to identify event relations, both within and across sentences. It's important to note that both of these approaches are pairwise methods and do not incorporate cross-document entity coreference into their methodologies. In contrast, our approach with X-AMR not only leverages cross-document entity coreference but also capitalizes on AMR's nested event structure for ECR.


\newcommand{\mI}{\ensuremath{m_i}}
\newcommand{\mJ}{\ensuremath{m_j}}
\newcommand{\corefm}{\ensuremath{\mathtt{coref}(\mI, \mJ)}}
\subsection{\eid: Event Identifiers}
We generate \eid~using the roleset, \argzero, and \argone. To evaluate the influence of location and time, we produce \eidLT~by incorporating \argL~and \argT. These identifiers facilitate comparison between two events, allowing coreference resolution by matching the identifiers. Specifically, two events ($m_i$, $m_j$) are deemed coreferent (where \corefm~is true) if any of their identifiers match in $\eid(m_i)$~and $\eid(m_j)$~match, as illustrated in Equation  \ref{eqn:coref-algo}.
\begin{equation}
\label{eqn:coref-algo}
\resizebox{\textwidth/2 - 35pt}{!}{
\ensuremath{\begin{split}
    \mathtt{coref}(m_i, m_j) \equiv \eid(m_i) \cap \eid(m_j) \neq \text{\O}
\end{split}
}}
\end{equation}
\noindent Even though Eq \ref{eqn:coref-algo} is represented pairwise, we design the clustering algorithm by first creating buckets of mentions with the same identifiers. This way we generate a \textit{sparse binary similarity matrix} of only the pairs of mentions in the same buckets representing the \eid s.
\newcommand{\msube}{\ensuremath{m_{e}}}
\newcommand{\mstandard}{\ensuremath{m}}
\subsubsection{\eid~Generation}
We generate the identifiers differently for Standard events (\mstandard, like \mone, \mtwo, \mthree) and Nested Events (\msube, such as \mfour~and \mfive). For standard events, the identifier (\eidzerohop) is constructed by merging the ARG-0, roleset ID (\RS), and ARG-1 as shown in Equation \ref{eqn:eidn}. For instance, \eid(\mone) is denoted as \textlangle HP, acquire.01, EYP\textrangle.
\newcommand{\RSVN}{\ensuremath{\RSHum_\PBSynVN}}
\newcommand{\RSGVN}{\ensuremath{\PB_{\mathtt{GVN}}}}
\newcommand{\RSHUM}{\ensuremath{\RSHum_\PBSynHum}}
\begin{equation}
\label{eqn:eidn}
\resizebox{\textwidth/2 - 35pt}{!}{
    \eidzerohop(\mstandard) = \text{\textlangle} \argzero(\mstandard),~\RS(\mstandard),~\argone(\mstandard)\text{\textrangle}
}
\end{equation}
\begin{figure}[t]
    \centering
    \resizebox{0.9\linewidth}{!}{\includegraphics{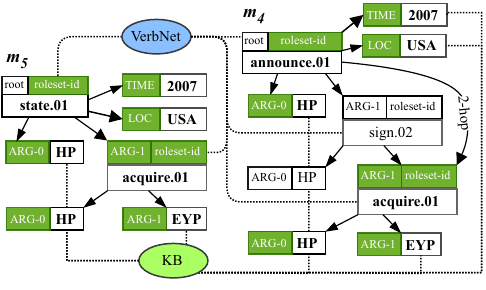}}
    \caption{Event Identifier and Coreference for events with eventive \argone. The \eidtwohop(\mfour) is equivalent \eidonehop(\mfive) with the help of VerbNet to detect synonymy and KB to link arguments.}
    \label{fig:coref_eg_eventive}
\end{figure}

\vspace*{-3mm}
In the case of Nested Events (ARG-1 is also an event), we employ a recursive strategy to generate identifiers. Specifically, we produce multiple \eid s by traversing the arguments of nested events up to a maximum depth,  N, as delineated in Equations \ref{eqn:eid} and \ref{eqn:eidhop}. This method aims to connect the root event to a standard ARG-1 within the event chain. This procedure is denoted as \eidnHop, where $n$ indicates the utilized depth. Notably, for our experiments, we set $n = \text{N}$ during \eid~generation.
\begin{equation}
\label{eqn:eid}
    \resizebox{\textwidth/2 - 40pt}{!}{
\ensuremath{
\begin{split}
    \eidnHop(\msube) = \bigcup_{k}^{n}
    \Big[
        \text{\textlangle} \argzero(&\msube) \concat  \RS(\msube) \text{\textrangle} \times \\
         & \eidhop(\argone(\msube))~
    \Big],
\end{split}
}
}
\end{equation}
where $\times$ is the Cartesian product for generating all the concatenations of the tuples.

\begin{align}
\label{eqn:eidhop}
\resizebox{\textwidth/2 - 35pt}{!}{
\ensuremath{
\eidhop(m) = 
\begin{cases}
\argone(m)  & \text{if standard},\\
& \hspace{2mm} \text{or}~k = n, \\
\eidnminusoneHop(\argone(m)) &  \text{otherwise}
\end{cases}
}
}
\end{align}

Using Eq \ref{eqn:eid} and \ref{eqn:eidhop}, we generate the identifiers for \mfour~and \mfive~(Figure \ref{fig:coref_eg_eventive}) as shown below: 

\begin{table}[h!]
    \centering
\resizebox{\columnwidth}{!}{
    \begin{tabular}{rl}
        \eidtwohop(\mfour) & \makecell[l]{\textlangle HP, announce.01, HP, sign.02, HP, acquire.01, EYP\textrangle, \\ \textlangle HP, announce.01, HP, acquire.01, EYP \textrangle}  \\ \midrule
        \eidonehop(\mfive)& \textlangle HP, state.01, HP, acquire.01, EYP \textrangle
    \end{tabular}
}
    \label{tab:eidns}
\end{table}
\noindent The 2-hop identifier of \mfour~is exactly same as the 1-hop one of \mfive, except for the roleset IDs. To detect synonymy between the rolesets, we use VerbNet (\RSVN), and we maintain a KB to link the arguments. With the combination of all these components, we can infer that \mfour~is coreferent with \mfive.

We also use the \argL~and \argT~to separately generate an identifier, \eidLT, for both kinds of events as shown in Equation \ref{eqn:eidLT}.
\begin{equation}
\label{eqn:eidLT}
\resizebox{\textwidth/2 - 60pt}{!}{
\ensuremath{
\begin{split}
    \eidLT (m) =~& \text{\textlangle} \argzero(m),~  \RS(m), \\ &~~\argL(m),~\argT(m) \text{\textrangle}
\end{split}
}
}
\end{equation}




\newcommand{\annone}{\ensuremath{\text{A}_{1}}}
\newcommand{\anntwo}{\ensuremath{\text{A}_{2}}}
\newcommand{\annOneAndTwo}{\ensuremath{\annone \wedge \anntwo}}
\newcommand{\annOneOrTwo}{\ensuremath{\annone \lor \anntwo}}
\newcommand{\AND}{\ensuremath{\wedge}}
\newcommand{\OR}{\ensuremath{\lor}}
\subsection{Clustering Methods}
\label{sec:clustering_methods}
We generate the adjacency matrix of the mentions by using certain baselines and the event identifiers. The adjacency matrix is then used for hard-clustering the events by finding the connected components.
\newcommand{\LEM}{\ensuremath{\mathtt{LEM}}}

\begin{itemize}[label={},leftmargin=1.5mm, itemsep=-0.5mm]
\item \textbf{Baseline (\LEM):} Clustering mentions with the same lemma for their triggers serves as our baseline method.
\item \textbf{Rolesets IDs (\RSHum, \RSGPT):} We cluster mentions based on the strict similarity of these PropBank Roleset IDs. We use the human (\RSHum) and GPT-generated (\RSGPT, see \S\ref{sec:gpt}) roleset IDs separately.
\item \textbf{RS with VerbNet Syn classes (\RSVN, 
 \RSGVN):} We cluster mentions of synonymous rolesets based on VerbNet Classes \cite{brown-etal-2011-verbnet, brown2022semantic} allowing less strict roleset matching.
\item \textbf{Event Identifiers}: We vary the \eid~methods in the following ways:
\begin{itemize}[itemsep=1.5mm, label={},leftmargin=1.5mm]
    \item \eidNH: We cluster with \eidNH~using \RSHum~or \RSGPT
    \item \eidLTH: We cluster with \eidLT~using \RSHum~or \RSGPT
    \item \eidHVN: \eidNH~only, but with \RSVN~or \RSGVN~to identify roleset classes for \RSHum~or \RSGPT
    \item \eidLTHVN: \eidLT~only, but with \RSVN~or \RSGVN~to identify roleset classes for \RSHum~or \RSGPT
    \item \eidNH~\AND~\eidLT: We cluster the mentions when they have the same \eidNH~and \eidLT.
    \item \eidNH~\OR~\eidLT: We cluster the mentions when they have either the same \eidNH~or \eidLT.
\end{itemize}
We also include the VerbNet class versions of the final two methods.
\end{itemize}

In addition to the individual methods listed above, we also employ combinations of methods on the annotations of the two annotators (\annone, \anntwo). $\land$-clustering employs the rule that two mentions should have the same annotations from \annone~and \anntwo~(\annOneAndTwo). $\lor$-clustering employs the rule that any of the two annotators's annotations could be the same for two mentions (\annOneOrTwo).

\section{ECR Results of \annone~and \anntwo}

We use the standard clustering metrics for ECR (MUC, $B^{3}$, $CEAF_e$, and CoNLL F1---the average of MUC, $B^{3}$ and $CEAF_e$; \citet{10.3115/1072399.1072405,Bagga98algorithmsfor, 10.3115/1220575.1220579, 515751742011, luo-etal-2014-extension, pradhan-EtAl:2014:P14-2,moosavi-etal-2019-using}). To evaluate recall, we compute the mean recall values from MUC and $B^{3}$ (R$_{avg}$). Similarly, our precision metric, P$_{avg}$ is derived from the average precision values of MUC and $B^{3}$. Our primary measure of overall performance is CoNLL F1. We applied various algorithmic methods to the ECB+ development set, which has been annotated using the X-AMR framework by \annone~and \anntwo. Each annotator's performance is independently assessed, along with the \OR-clustering approach, \annOneOrTwo. We collate the results in Table \ref{tab:ecb_res_ann}\footnote{\annOneAndTwo~results are excluded due to inferior quality.}.

\newcommand{\conll}{\small CoNLL}
 \renewcommand{\arraystretch}{1.3}
\begin{table}[t]
\centering
\small
    \begin{tabular}{@{}c|lcccc@{}}
        \toprule
       \multicolumn{1}{c}{~}& Method && R$_{avg}$ & P$_{avg}$ & \conll \\ 
        \midrule
        \multicolumn{1}{c}{~}&\LEM && 72.6 & 64.0 & 63.7  \\
        \multicolumn{1}{c}{~}&\RSHum && 81.2 & 63.5 & 66.1 \\
        \multicolumn{1}{c}{~}& \RSVN && \textbf{91.0} & 43.0 & 44.9 \\
 \hline
 \multirow{5}{*}{\rotatebox{90}{\annone}}
& \eidNH && 75.6 & 90.5 & 78.4 \\
& \eidLTH && 77.9 & 91.2 & \textbf{79.8} \\
& \eidNH~\AND~\eidLTH && 74.2 & \textbf{92.8} & 78.4 \\
& \eidNH~\OR~\eidLTH && 79.3 & 88.9 & \textbf{79.8} \\
& \eidHVN~\AND~\eidLTHVN && 80.0 & 84.2 & 78.5 \\
\hline
  \multirow{4}{*}{\rotatebox{90}{\anntwo}}
& \eidNH && 69.8 & 89.4 & 75.0 \\
& \eidLTH && 66.6 & 88.8 & 72.1 \\
& \eidNH~\AND~\eidLTH && 61.0 & 91.8 & 69.7 \\
& \eidNH~\OR~\eidLTH && 76.0 & 86.4 & 77.1 \\
\hline
\multirow{4}{*}{\rotatebox{90}{\annOneOrTwo}}
& \eidNH && 79.0 & 82.9 & 77.4 \\   
& \eidLTH && 80.2 & 83.6 & 77.8 \\
& \eidNH~\AND~\eidLTH && 78.4 & 85.6 & 78.3 \\
& \eidNH~\OR~\eidLTH && 80.8 & 80.8 & 76.9 \\
& \eidHVN~\AND~\eidLTHVN && \textbf{86.9} & 69.1 & 73.1 \\
\hline
\bottomrule
    \end{tabular}
  \caption[\ecb~Results]{
	ECR results comparing the annotators on the Development Set of the ECB+ Corpus. We report the baseline results using only the lexical information, and, the ECR performance of the proposed graph-based algorithm on the X-AMR annotations of \annone~and \anntwo, and, a union of \annone~and \anntwo~(\annone \OR \anntwo). Boldened are the interesting results. }
\label{tab:ecb_res_ann}
\end{table}

From the table, it is evident that utilizing the roleset IDs (\RSHum) achieves a better result than lemmas (\LEM). Even though \RSVN~has the highest recall of 91\%, the overall performance is quite low. The 9\% recall error indicates the gap in the VerbNet class annotations for all the PropBank rolesets.  This suggests there may be room for refining the VerbNet-Pro  annotations for better compatibility \cite{spaulding-etal-2024-propbank}.

Comparing annotators, \annone~provided more accurate annotations than \anntwo, particularly in identifying location and time elements.  Both annotators performed best with the \eidNH~\OR~\eidLT~setting, with \annone~recording the best CoNLL F1 score of 79.8\%. \anntwo's annotations would need further refinement in order to match \annone's recall. When considering precision, the \eidNH~\AND~\eidLT~method stood out, with \annone registering the highest precision at 92.8\%. 

For \annOneOrTwo, the results are mixed. Although the recall is consistently higher than any individual annotator, it does not beat \annone's best CoNLL. This method achieves the best recall of 86.9 when used in conjunction with the VerbNet classes while also having a CoNLL F1 greater than 70\%. The mixed results for the combined method underscore the complexities involved in integrating and harmonizing annotations from different sources.

A CoNLL F1 of 80\% seems to be an upper bound for a purely symbolic approach for ECR. However, we want to stress that after annotating X-AMR, we are in effect collecting free ECR annotations (eg. 75\% coreference links with 93\% precision). ECR annotations are traditionally done in a pairwise manner, an approach that is tedious and error-prone \cite{song-etal-2018-cross, wright-bettner-etal-2019-cross}. In contrast, X-AMR has an annotator-friendly methodology where an annotator would need to read a particular event mention typically only once. It also avoids annotation errors cascading into subsequent mentions as demonstrated by the high precision of our method.
\section{GPT-4 as Annotator}
\label{sec:gpt}
\newcommand{\smallDev}{\ensuremath{\text{dev}_{\text{small}}}}
\newcommand{\gOne}{\ensuremath{\text{G}_{1}}}
\newcommand{\gTwo}{\ensuremath{\text{G}_{2}}}
Recent work in prompt engineering converts a text-based natural language task to a corresponding structured prediction task. In this spirit, we create prompts for extracting the X-AMR graph for a specific event, by providing the instructions for the task, exemplars for the structure of the response, and the right context for in-context zero-shot learning. Due to budget and time constraints, we make a smaller subset (120 mentions) of the development set (\smallDev) to run our experiments on. We then assess the performance of GPT-4 (September 27, 2023 version) against the human annotations for this subset. We try two prompt engineering techniques (\gOne, \gTwo) to extract the X-AMR graphs of the events, and use the \eid~generation and clustering methods from \S \ref{sec:clustering_methods}.

\subsection{\gOne: Prompt Engineering}
For \gOne, we use a straightforward approach to generate the prompts. We start by generating a list of instructions. As shown in Figure \ref{fig:prompt_instructions}, we arrive at five instructions by condensing the relevant sections of the annotation guidelines. We adopt a semi-automated way of generating the instructions, in which we first pass the relevant sections to ChatGPT and then hand-correct its output.

\subsubsection{Structured Prediction: Label Definitions}
We then prompt GPT to produce a JSON output as the response. We offer detailed definitions for the keys in the JSON string, as illustrated in Figure \ref{fig:key_defs}. Additionally, we incorporate the coreference key and prompt GPT to generate Wikipedia links in the format ``/wiki/Title\_Name". Labels for Chain of Thought reasoning \cite{NEURIPS2022_9d560961} are also included, addressing questions like ``Is it a Nested Event?", ``What is the event trigger?", ``Who are the participants?", and ``When and where did the event take place?". The final key in the list is ``Event Description" that is a way to prompt GPT to produce a concise sentence encapsulating the event arguments including Time and Location.

Finally, we add the entire document of the event and the sentence with the marked trigger (phrase in the sentence sorrounded by <m> and </m>) as context, and then prompt GPT to generate the corresponding JSON response.
\begin{figure}[t!]
\begin{mybox3}{\helvetexttitle{Label Definitions}}
\boxsize
\hspace{-1pt}Here are the definitions of the keys in the JSON output:
\vspace*{-1.5mm}
\begin{itemize}[label={}, leftmargin=1pt]
\setlength{\itemsep}{1pt}
  \setlength{\parskip}{0.3pt}
    \item \textbf{Roleset ID}: The PropBank Roleset ID corresponding to the event trigger
    \item \textbf{ARG-0}: The text in the Document corresponding to the typical agent
    \item \textbf{ARG-0 Coreference}: The reference to the ARG-0 in Wikipedia in the format /wiki/Wikipedia\_ID
    \item $\vdots$
    \item \textbf{ARG-1 Roleset ID}: If the Event is Nested, provide the Roleset ID for the head event in ARG-1 clause
    \item \textbf{ARG-Location}: The reference to the event location in Wikipedia
    \item \textbf{ARG-Time}: The event time in the format of Month-Day-Year in your knowledge of the world or the document
    \item \textbf{Event Description}: In a single sentence, summarize the event capturing the Roleset\_ID and the names and wiki links of the Participants, Location and Time
\end{itemize}
\end{mybox3}
\caption{Label definitions for the event's Roleset ID and the Arguments that include the Wikipedia links. Event Description is a single sentence encapsulating the key components of the event.}
    \label{fig:key_defs}
\end{figure}

\begin{figure}[t]
\begin{mybox3}{\helvetexttitle{Annotation Instructions}}
\boxsize
\hspace{-1pt}You are a concise annotator that follows these instructions:
\vspace*{-1.5mm}
\begin{enumerate}[leftmargin=*]
\setlength{\itemsep}{2pt}
  \setlength{\parskip}{0.5pt}
    \item Identify the target event trigger lemma and its correct roleset sense in the given text.
    \item Annotate the document-level \argzero\space and \argone\space roles using the PropBank website for the roleset definitions.
    \item If the \argone\space role is an event, identify the head predicate and provide its roleset ID.
    \item Perform within-document and cross-document ana\-phora resolution of the \argzero\space and \argone\space using Wikipedia.
    \item Use external resources, such as Wikipedia, to annotate \argL\space and \argT .
\end{enumerate}
\end{mybox3}
\caption{The condensed annotation instructions serve as a guide for GPT-4 in its generation of X-AMR event extraction.}
    \label{fig:prompt_instructions}
\vspace*{-2.5mm}
\end{figure}
\vspace{-1mm}
\subsection{\gTwo: Prompt Engineering}
A challenge observed in \gOne~ is its inability to determine specific pieces of information, particularly, `Location' and `Time' when they are absent within the source document. To address this shortfall, we introduce a complementary method, \gTwo.

\begin{itemize}[label={}, leftmargin=1.5mm]
\setlength{\itemsep}{1.5pt}
    \item \textbf{Event Descriptions:} In \gTwo, instead of relying solely on the document's raw content, we incorporate additional context derived from the event descriptions of what we term as \textit{complete} events. These complete events are identified across all documents related to a specific topic at the prediction stage. A \textit{complete} event is characterized by having all its requisite arguments, including Time and Location, predicted by \gOne.
    
    \item \textbf{De-duplication:} To enhance the quality and relevancy of this list, any coreferent events (duplicates) are eliminated.
    
    \item \textbf{Event List in Context:} With the refined list, we pivot from using the entire document as context (as practiced in \gOne) to utilizing this labeled list of Event Descriptions. Furthermore, the description of the current target event is also included.
    
    \item \textbf{Best Matching Event Description:} We introduce a label called ``Best Matching Event Description" at the beginning of prediction. This label pinpoints the most comprehensive and relevant description in correlation to the target mention. The intention behind this is to direct GPT's attention to a singular event description, enabling it to supplement the arguments not identified by \gOne.
\end{itemize}

In essence, \gTwo~furnishes a richer context, combining aggregated information from various documents, to rectify the limitations observed in \gOne.

\section{ECR Results of \gOne~and \gTwo}
We compare the methods \gOne~and \gTwo~(The cost for running \gOne~was \$4, and \gTwo~was \$6.) separately with \annone~(the annotations with better quality among the two annotators) on \smallDev. As shown in Table \ref{tab:ecb_res_gpt}, the roleset identification by GPT-4 is impressive, thereby we only see a 3 point difference between \RSHum~and \RSGPT. In \smallDev, we observe the results are bounded by recall, therefore we use the VerbNet class approaches.

\renewcommand{\arraystretch}{1.3}
\begin{table}[t!]
\centering
\small
    \begin{tabular}{@{}c|lcccc@{}}
        \toprule
       \multicolumn{1}{c}{~}& Method && R$_{avg}$ & P$_{avg}$ & \conll \\ 
        \midrule
        \multicolumn{1}{c}{~}&\LEM && 57.2 & 84.8 & 65.1 \\ 
        \multicolumn{1}{c}{~}&\RSHum && 72.2 & 85.7 & \textbf{75.3} \\
        \multicolumn{1}{c}{~} & \RSGPT && 70.6 & 80.6 & 72.4 \\
        \multicolumn{1}{c}{~}&  \RSVN && \textbf{90.4} & 51.8 & 55.9 \\
        \multicolumn{1}{c}{~} & \RSGVN && 87.6 & 46.7 & 49.3 \\

 \hline
 \multirow{3}{*}{\rotatebox{90}{\annone}}
& \eidNH && 68.8 & \textbf{100} & 77.7 \\
& \eidHVN && 78.4 & 97.2 & \textbf{83.6} \\
& \eidHVN~\OR~\eidLTHVN && 80.8 & 93.3 & 83.4 \\
\hline
  \multirow{3}{*}{\rotatebox{90}{\gOne}}
& \eidHVN && 41.8 & 88.9 & 53.6 \\
& \eidLTHVN && 37.2 & 83.0 & 49.4 \\
& \eidHVN~\OR~\eidLTHVN && 51.2 & 83.8 & 58.0 \\

\hline
\multirow{3}{*}{\rotatebox{90}{\gTwo}}
& \eidHVN && 51.8 & 90.6 & 62.2 \\
& \eidLTHVN && 57.0 & 87.4 & 65.3 \\
& \eidHVN~\OR~\eidLTHVN && 63.4 & 86.1 & \textbf{68.4} \\
\hline
\bottomrule
    \end{tabular}
  \caption[\ecb~Results]{
	ECR results comparing \annone~with the two prompting methods \gOne~and \gTwo, on \smallDev. We report the baseline results using only the lexical information, and, the ECR performance after leveraging X-AMR.  Boldened are the interesting results. }
\label{tab:ecb_res_gpt}
\end{table}

When using \eid~methods, \annone~achieves the best CoNLL F1 of 83.6. When it comes to GPT-4, both \gOne~and \gTwo, fell terribly short of \annone~and do not even surpass the Roleset ID baseline (\RSGPT), with \gOne's best performance is short by 25 points and \gTwo~by 15. The shortcoming of \gTwo~can mainly be attributed to the failure of capturing nested events (only 5 of the 26 nested event arguments were identified). Interestingly, these methods consistently improve performance over the VerbNet baseline (\RSGVN). Between \gOne~and \gTwo, we see a large performance increase (\gTwo~over \gOne~by 10 points), emphasizing the benefits of using corpus-level Event Descriptions in the prompts. 

The results reveal the limitations of GPT-4 on this task. However, efficient usage of corpus-level information in generating X-AMR graphs lays out an exciting path forward for future work.
\vspace*{-5.5mm}
\section{Analysis}
\subsection{Algorithm Complexity}
\begin{figure}[b!]
    \centering
    \vspace*{-1.5mm}
    \includegraphics[scale=0.5]{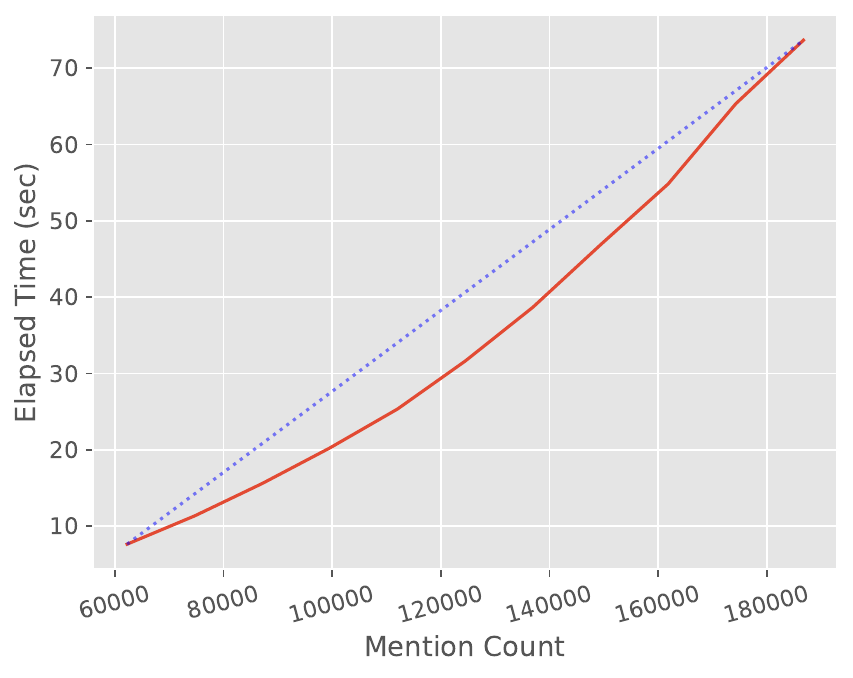}
    \vspace*{-1.5mm}
    \caption{X-AMR ECR Algorithm running time on synthetic mentions. The dotted line is used as a reference to check the linearity of the algorithm (red line).The complexity is roughly linear.}
    \label{fig:complexity}
\end{figure}

\begin{table*}[t!]
\centering%
\small
\begin{tabularx}{\linewidth}{llX}
\toprule
{\small Error Category} & {\small \% Error} & {\small Snippet and Explanation} \\
\midrule
\midrule

{\small \multirow{3}{*}{Annotation Error}} & \multirow{3}{*}{1.6} & \mone: \underline{The man} is thought to have \textbf{fallen} much earlier in the day.\\
& & \mtwo: [...] \underline{Duncan Rait} died after slipping and  \textbf{falling} [...]\\
& & \small Explanation: annotator misidentifies"the man" as "Duncan Rait". \\
\midrule

{\small \multirow{3}{*}{ECB+ Annotation Error}} & \multirow{3}{*}{5.8} & \mone: [...] the \underline{Philly Sixers} \textbf{canned} Jim O'Brien [...] \\
& & \mtwo: Jim O'Brien was \textbf{terminated} from [...] \underline{Ohio State University} [...]\\ & & \small Explanation: In ECB+ these events are falsely labelled as coreferent.\\
\midrule

{\small \multirow{2}{*}{Incorrect VerbNet Class}} & \multirow{2}{*}{9.1} & [...] who was  \textbf{\underline{gunned down}}  at their office Christmas party. \\
& & \small Explanation: new Propbank rolesets aren't yet mapped to VerbNet.\\
\bottomrule
\end{tabularx}
\caption {Qualitative Error Analysis of the ECR Algorithm for \annone~on \smallDev. In Snippet, the event triggers are in \textbf{bold font}, and the key texts that help recognize the errors are \underline{underlined}.}
    \label{tab:error_table_full}
\end{table*}

We conduct an artificial experiment to empirically demonstrate the ECR algorithm's linear complexity when coupled with X-AMR. In this experiment, we expand the annotated event mentions of the development set (by duplicating) to create a sizable collection comprising 200,000 mentions. Next, we systematically execute the algorithm across varying ranges, from 60,000 to 200,000 mentions. For each iteration, we measure the time the ECR algorithm takes to run and depict the plot in Figure \ref{fig:complexity}. The figure shows that the algorithm's running time is roughly linear. Efficiency-wise, the algorithm would take under 30 seconds even when the number of mentions surpasses 100,000, thus presenting a tractable solution for ECR at scale.

\subsection{Error Analysis}
We analyze the errors made by our system by examining the clustering decisions based on the \annone, \gOne, and \gTwo~annotations of \smallDev~(121 mentions).

In the human-annotated \annone, we observe that $\sim$1.6\% (2/121) of the misclassifications were due to annotator misinterpretations of the passages. In another $\sim$5.8\% (7/121) of cases, an incorrect cluster is assigned due to errors in the original ECB+ dataset's labels, which are made evident by mismatched X-AMR arguments. For example, the ECB+ labels erroneously consider football coach Jim O'Brien's separate terminations from Ohio State University and the Philadelphia 76ers as the same event. Finally, $\sim$9.1\% (11/121) are misclustered due to PropBank labels which do not yet exist in VerbNet and so do not belong to a class; e.g. the newer ``opening fire" roleset wasn't identified as being of the same class as ``shoot". Examples of each type of error are provided in Table \ref{tab:error_table_full}.

In addition to the problems faced by human annotators, the machine-created annotations \gOne~and \gTwo~suffered heavily from their inability to access external resources to resolve relative times, locations, and references to entities, in addition to inconsistent annotations (which the human annotators did not suffer from due to the saved argument drop-down), e.g ['South\_Richmond\_Hill,\_Queens', 'Queens', 'Richmond\_Hill,\_Queens'] all refer to the same place\footnote{For a more comprehensive list of examples, please refer to the provided Excel file in the repository}.
\section{Limitations \& Future Work}
One limitation of our approach is that we require the PropBank resource for a particular language. In addition, the annotation tool is for-pay software. However, PropBank now has annotations for Chinese, Arabic, Urdu,
Hindi, French, German, Spanish, and Catalan, and the annotation tool also works on a variety of languages.  Our future work involves annotating X-AMR on a Spanish corpus. 

The annotation tool released by \citet{ahmed-etal-2024-xtool} omits a lot of AMR information (e.g. modality and negation), sticking strictly to the concept of minimal information for ECR. We also do not empirically demonstrate the efficiency of the model-in-the-loop annotations in this work. We leave the tool enhancements, including the incorporation of GPT-in-the-loop and a thorough analysis of the annotation efficiency (like \citet{jon-etal-2023-camra} and \citet{ahmed-etal-2023-good}) for future work.

The results for both human and GPT-annotated approaches fall short of state-of-the-art techniques for ECR that involve heuristics (for filtering) and fine-tuning BERT in a pairwise manner \cite{held-etal-2021-focus, ahmed-etal-2023-2}. We hypothesize that the X-AMR annotations might be beneficial to the heuristic-based filtering step in these methods. The Event Description generated by \gOne~could also be employed while fine-tuning BERT which we believe is an interesting direction for neuro-symbolic methods for ECR.

Two main issues of using GPT-4 in our work are Data contamination \cite{magar-schwartz-2022-data, wu2023reasoning} and reproducibility. Since both PropBank and ECB+ are publicly available resources, it is most likely that the test sets might be part of its pre-training data. We argue that since our task is vastly different from the pretraining task, the effect of contamination is minute as demonstrated by the results. Reproducibility, however, is a much greater limitation. By providing the GPT-4 output on the train set (will release this upon acceptance), we set a mechanism to distill the knowledge into smaller in-house reproducible models like LlAma \cite{touvron2023llama} (Or even much smaller traditional auto-regressive models like FLAN-T5 \cite{flan-t5}) for future work.

Finally, we limit the scope of our work to gold mentions instead of predicted mentions \cite{cattan-etal-2021-cross-document}. As a result, we could not compare X-AMR directly with the output of standard AMR parsers \cite{flanigan-etal-2014-discriminative}. Future work can approach this in a two-step way, with the first step being trigger identification, and then we can employ X-AMR on the predicted mentions.
%



\vspace{-1.3mm}
\section{Related Work}
\vspace{-0.6mm}


Document-level event extraction and event extraction with prompts \cite{li-etal-2021-document, Yang2022FewShotDE, xu-etal-2022-two} has been a major source of inspiration for our work. We extend this methodology for a more comprehensive cross-document level extraction by taking into account the named and unnamed arguments from previously seen documents into the annotation framework.

The Generative Pre-trained Transformer (GPT; \citet{radford2018improving, radford2019language}) is an auto-regressive Transformer \cite{NIPS2017_3f5ee243} language model developed by OpenAI, demonstrating exceptional performance across various natural language processing tasks. It uses a unidirectional, self-attention mechanism for effective context representation and is pre-trained on extensive unsupervised text corpora. The model follows a two-stage process of pre-training and fine-tuning, allowing it to adapt to specific tasks with minimal labeled data. GPT has undergone several iterations, with GPT-4 \cite{openai2023gpt4} being the most recent.

In recent years, research has increasingly focused on evaluating GPT's performance in multi-task and zero/few-shot learning scenarios \cite{NEURIPS2020_1457c0d6, kojima2023large}. For instance, the study conducted by \citet{radford2019language} assesses the effectiveness of various LLMs in a zero-shot learning setting. Their findings imply that these models have the potential to equal, if not exceed, the performance of existing baselines on a range of NLP benchmarks.



Our objective is to underscore the importance of X-AMR with a focus on event coreference resolution, which integrates PropBank \cite{palmer-etal-2005-proposition, pradhan-etal-2022-propbank} SRL as an intermediate phase. This approach is motivated by GPT-4's capability to produce free-text SRL for individual events \cite{zhang-etal-2022-transfer} instead of directly generating interconnected event graphs, as necessitated by AMR. By leveraging GPT-4's strengths, our suggested method can offer a more thorough and effective representation of events in a given text while preserving their structure and relationships, and therefore facilitate ECR.


Besides the hybrid approach and prompt engineering, we also stress the need for a linear algorithm over a quadratic ECR method, utilizing the generated graphs. Quadratic ECR with GPT (i.e., binary coreference decision between mention pairs) has produced negative outcomes, as evidenced by \citet{yang-etal-2022-gpt}. Furthermore, this method would be expensive, potentially costing hundreds of dollars to execute using GPT-4. By adopting a linear algorithm, we aim to address these limitations, offering a more cost-effective and efficient solution for ECR. We propose a linear graph-based method for ECR using the generated key semantic information for the event mentions.

Over time, efforts have been made to enrich event datasets, such as the Richer Event Descriptions (RED; \citet{ogorman-etal-2016-richer}) corpus and the Event Coref Bank plus corpus (ECB+; \citet{cybulska-vossen-2014-using}). The RED corpus enhanced ERE \cite{song-etal-2015-light} annotations by marking coreference for entities, events, and times, as well as temporal, causal, and subevent relationships in partial coreference through a multi-stage pipeline. In ECB+, \citet{cybulska-vossen-2014-using} expanded event descriptions by adding event classes with specific entity types and times, as well as inter-/intra-document coreference, to better represent the events within the ECB corpus \cite{bejan-harabagiu-2010-unsupervised}. In a similar light, we enrich the ECB+ corpus with the X-AMR annotations with the goal of making ECR efficient and as a way to assess the performance of GPT-4.

\vspace{-1mm}
\section{Conclusion}
\vspace{-0.6mm}
In this paper, we introduced X-AMR, a corpus-level version of AMR. We provided a new model-in-the-loop tool with which we enriched the ECB+ corpus with X-AMR annotations. We then introduced a novel linear graph-based ECR algorithm that leverages the nested event structure and the cross-document entity coreference of X-AMR. The annotations coupled with the algorithm serve as a way for linearly generating cross-document event coreference annotations, cutting through a very challenging task. Finally, we developed two prompt engineering approaches for GPT-4 to automatically produce X-AMR graphs. We then compared the results against human annotations and showed limitations of GPT-4 on this task. We also provide comprehensive and concise GPT-generated event descriptors in this process that we believe have a lot of utility in other event tasks. Collectively, our contributions pave a path toward efficient ECR methods and their corresponding annotations.





\section*{Ethics Statement}
Recognizing the rigor and tediousness of the annotation process, our research ensured that all annotators were fairly compensated, given reasonable work hours, and provided with regular breaks to maintain consistency and quality. Comprehensive training and clear guidelines were offered, and a robust communication channel was established to address concerns, ambiguities, and to encourage feedback. Our team made efforts to involve a diverse group of annotators to minimize biases.

To alleviate the monotonous nature of the task, we employed user-friendly tools, rotated tasks, and supported peer discussions. We also acknowledged the crucial role of annotators in our research, ensuring their contributions were recognized and valued. Post-task, a summary of our findings was shared with the annotators, incorporating their feedback into the final manuscript, underlining our commitment to an inclusive and ethical research approach.

By adhering to the LREC-COLING guidelines, we aim to emphasize the ethical considerations surrounding the involvement of annotators in research projects. We believe that a humane, respectful, and inclusive approach to data annotation not only results in superior-quality datasets but also upholds the dignity and rights of all involved.
\section*{Acknowledgements}
We want to thank the reviewers of LREC-COLING 2024 who helped improve this paper. Part of this work was done during an internship of one of the authors at ExplosionAI GmbH. We would also like to thank Ákos Kádár, Matthew Hannibal, Nikhil Krishnaswamy, Elizabeth Spaulding, and the BoulderNLP group for their valuable comments on this paper. We gratefully acknowledge the support of  DARPA FA8750-18-2-0016-AIDA – RAMFIS: Representations of vectors and Abstract Meanings for Information Synthesis and a sub-award from RPI on DARPA KAIROS Program No. FA8750-19-2-1004.  Any opinions, findings, conclusions, or recommendations expressed in this material are those of the authors and do not necessarily reflect the views of DARPA or the U.S. government.
\section{Bibliographical Resources}
\bibliography{anthology,custom}

\begin{thebibliography}{54}
\expandafter\ifx\csname natexlab\endcsname\relax\def\natexlab#1{#1}\fi

\bibitem[{Ahmed et~al.(2024)Ahmed, Cai, Palmer, and
  Martin}]{ahmed-etal-2024-xtool}
Shafiuddin~Rehan Ahmed, Jon Cai, Martha Palmer, and James~H. Martin. 2024.
\newblock \href {https://aclanthology.org/2024.eacl-demo.19} {{X}-{AMR}
  annotation tool}.
\newblock In \emph{Proceedings of the 18th Conference of the European Chapter
  of the Association for Computational Linguistics: System Demonstrations},
  pages 177--186, St. Julians, Malta. Association for Computational
  Linguistics.

\bibitem[{Ahmed et~al.(2023{\natexlab{a}})Ahmed, Nath, Martin, and
  Krishnaswamy}]{ahmed-etal-2023-2}
Shafiuddin~Rehan Ahmed, Abhijnan Nath, James~H. Martin, and Nikhil
  Krishnaswamy. 2023{\natexlab{a}}.
\newblock \href {https://doi.org/10.18653/v1/2023.findings-acl.100} {$2*n$ is
  better than $n^2$: Decomposing event coreference resolution into two
  tractable problems}.
\newblock In \emph{Findings of the Association for Computational Linguistics:
  ACL 2023}, pages 1569--1583, Toronto, Canada. Association for Computational
  Linguistics.

\bibitem[{Ahmed et~al.(2023{\natexlab{b}})Ahmed, Nath, Regan, Pollins,
  Krishnaswamy, and Martin}]{ahmed-etal-2023-good}
Shafiuddin~Rehan Ahmed, Abhijnan Nath, Michael Regan, Adam Pollins, Nikhil
  Krishnaswamy, and James~H. Martin. 2023{\natexlab{b}}.
\newblock \href {https://doi.org/10.18653/v1/2023.law-1.14} {How good is the
  model in model-in-the-loop event coreference resolution annotation?}
\newblock In \emph{Proceedings of the 17th Linguistic Annotation Workshop
  (LAW-XVII)}, pages 136--145, Toronto, Canada. Association for Computational
  Linguistics.

\bibitem[{Bagga and Baldwin(1998)}]{Bagga98algorithmsfor}
Amit Bagga and Breck Baldwin. 1998.
\newblock \href
  {https://citeseerx.ist.psu.edu/document?repid=rep1&type=pdf&doi=ccdacc60d9d68dfc1f94e7c68bd56646c000e4ab}
  {Algorithms for scoring coreference chains}.
\newblock In \emph{In The First International Conference on Language Resources
  and Evaluation Workshop on Linguistics Coreference}, pages 563--566.

\bibitem[{Banarescu et~al.(2013)Banarescu, Bonial, Cai, Georgescu, Griffitt,
  Hermjakob, Knight, Koehn, Palmer, and
  Schneider}]{banarescu-etal-2013-abstract}
Laura Banarescu, Claire Bonial, Shu Cai, Madalina Georgescu, Kira Griffitt, Ulf
  Hermjakob, Kevin Knight, Philipp Koehn, Martha Palmer, and Nathan Schneider.
  2013.
\newblock \href {https://aclanthology.org/W13-2322} {{A}bstract {M}eaning
  {R}epresentation for sembanking}.
\newblock In \emph{Proceedings of the 7th Linguistic Annotation Workshop and
  Interoperability with Discourse}, pages 178--186, Sofia, Bulgaria.
  Association for Computational Linguistics.

\bibitem[{Bejan and Harabagiu(2010)}]{bejan-harabagiu-2010-unsupervised}
Cosmin Bejan and Sanda Harabagiu. 2010.
\newblock \href {https://aclanthology.org/P10-1143} {Unsupervised event
  coreference resolution with rich linguistic features}.
\newblock In \emph{Proceedings of the 48th Annual Meeting of the Association
  for Computational Linguistics}, pages 1412--1422, Uppsala, Sweden.
  Association for Computational Linguistics.

\bibitem[{Beltagy et~al.(2020)Beltagy, Peters, and
  Cohan}]{beltagy2020longformer}
Iz~Beltagy, Matthew~E. Peters, and Arman Cohan. 2020.
\newblock \href {http://arxiv.org/abs/2004.05150} {Longformer: The
  long-document transformer}.

\bibitem[{Brown et~al.(2022)Brown, Bonn, Kazeminejad, Zaenen, Pustejovsky, and
  Palmer}]{brown2022semantic}
Susan~Windisch Brown, Julia Bonn, Ghazaleh Kazeminejad, Annie Zaenen, James
  Pustejovsky, and Martha Palmer. 2022.
\newblock \href {https://doi.org/10.3389/frai.2022.821697} {Semantic
  representations for nlp using verbnet and the generative lexicon}.
\newblock \emph{Frontiers in Artificial Intelligence}, 5.

\bibitem[{Brown et~al.(2011)Brown, Dligach, and
  Palmer}]{brown-etal-2011-verbnet}
Susan~Windisch Brown, Dmitriy Dligach, and Martha Palmer. 2011.
\newblock \href {https://aclanthology.org/W11-0110} {{V}erb{N}et class
  assignment as a {WSD} task}.
\newblock In \emph{Proceedings of the Ninth International Conference on
  Computational Semantics ({IWCS} 2011)}.

\bibitem[{Brown et~al.(2020)Brown, Mann, Ryder, Subbiah, Kaplan, Dhariwal,
  Neelakantan, Shyam, Sastry, Askell, Agarwal, Herbert-Voss, Krueger, Henighan,
  Child, Ramesh, Ziegler, Wu, Winter, Hesse, Chen, Sigler, Litwin, Gray, Chess,
  Clark, Berner, McCandlish, Radford, Sutskever, and
  Amodei}]{NEURIPS2020_1457c0d6}
Tom Brown, Benjamin Mann, Nick Ryder, Melanie Subbiah, Jared~D Kaplan, Prafulla
  Dhariwal, Arvind Neelakantan, Pranav Shyam, Girish Sastry, Amanda Askell,
  Sandhini Agarwal, Ariel Herbert-Voss, Gretchen Krueger, Tom Henighan, Rewon
  Child, Aditya Ramesh, Daniel Ziegler, Jeffrey Wu, Clemens Winter, Chris
  Hesse, Mark Chen, Eric Sigler, Mateusz Litwin, Scott Gray, Benjamin Chess,
  Jack Clark, Christopher Berner, Sam McCandlish, Alec Radford, Ilya Sutskever,
  and Dario Amodei. 2020.
\newblock \href
  {https://proceedings.neurips.cc/paper_files/paper/2020/file/1457c0d6bfcb4967418bfb8ac142f64a-Paper.pdf}
  {Language models are few-shot learners}.
\newblock In \emph{Advances in Neural Information Processing Systems},
  volume~33, pages 1877--1901. Curran Associates, Inc.

\bibitem[{Caciularu et~al.(2021)Caciularu, Cohan, Beltagy, Peters, Cattan, and
  Dagan}]{caciularu-etal-2021-cdlm-cross}
Avi Caciularu, Arman Cohan, Iz~Beltagy, Matthew Peters, Arie Cattan, and Ido
  Dagan. 2021.
\newblock \href {https://doi.org/10.18653/v1/2021.findings-emnlp.225} {{CDLM}:
  Cross-document language modeling}.
\newblock In \emph{Findings of the Association for Computational Linguistics:
  EMNLP 2021}, pages 2648--2662, Punta Cana, Dominican Republic. Association
  for Computational Linguistics.

\bibitem[{Cai et~al.(2023)Cai, Ahmed, Bonn, Wright-Bettner, Palmer, and
  Martin}]{jon-etal-2023-camra}
Jon Cai, Shafiuddin~Rehan Ahmed, Julia Bonn, Kristin Wright-Bettner, Martha
  Palmer, and James~H. Martin. 2023.
\newblock \href {https://doi.org/10.18653/v1/2023.emnlp-demo.35} {{CAMRA}:
  Copilot for {AMR} annotation}.
\newblock In \emph{Proceedings of the 2023 Conference on Empirical Methods in
  Natural Language Processing: System Demonstrations}, pages 381--388,
  Singapore. Association for Computational Linguistics.

\bibitem[{Cattan et~al.(2021)Cattan, Eirew, Stanovsky, Joshi, and
  Dagan}]{cattan-etal-2021-cross-document}
Arie Cattan, Alon Eirew, Gabriel Stanovsky, Mandar Joshi, and Ido Dagan. 2021.
\newblock \href {https://doi.org/10.18653/v1/2021.findings-acl.453}
  {Cross-document coreference resolution over predicted mentions}.
\newblock In \emph{Findings of the Association for Computational Linguistics:
  ACL-IJCNLP 2021}, pages 5100--5107, Online. Association for Computational
  Linguistics.

\bibitem[{Choubey and Huang(2017)}]{choubey-huang-2017-event}
Prafulla~Kumar Choubey and Ruihong Huang. 2017.
\newblock \href {https://doi.org/10.18653/v1/D17-1226} {Event coreference
  resolution by iteratively unfolding inter-dependencies among events}.
\newblock In \emph{Proceedings of the 2017 Conference on Empirical Methods in
  Natural Language Processing}, pages 2124--2133, Copenhagen, Denmark.
  Association for Computational Linguistics.

\bibitem[{Chung et~al.(2022)Chung, Hou, Longpre, Zoph, Tay, Fedus, Li, Wang,
  Dehghani, Brahma, Webson, Gu, Dai, Suzgun, Chen, Chowdhery, Narang, Mishra,
  Yu, Zhao, Huang, Dai, Yu, Petrov, Chi, Dean, Devlin, Roberts, Zhou, Le, and
  Wei}]{flan-t5}
Hyung~Won Chung, Le~Hou, Shayne Longpre, Barret Zoph, Yi~Tay, William Fedus,
  Eric Li, Xuezhi Wang, Mostafa Dehghani, Siddhartha Brahma, Albert Webson,
  Shixiang~Shane Gu, Zhuyun Dai, Mirac Suzgun, Xinyun Chen, Aakanksha
  Chowdhery, Sharan Narang, Gaurav Mishra, Adams Yu, Vincent Zhao, Yanping
  Huang, Andrew Dai, Hongkun Yu, Slav Petrov, Ed~H. Chi, Jeff Dean, Jacob
  Devlin, Adam Roberts, Denny Zhou, Quoc~V. Le, and Jason Wei. 2022.
\newblock \href {https://doi.org/10.48550/ARXIV.2210.11416} {Scaling
  instruction-finetuned language models}.

\bibitem[{Cybulska and Vossen(2013)}]{cybulska-vossen-2013-semantic}
Agata Cybulska and Piek Vossen. 2013.
\newblock \href {https://aclanthology.org/R13-1021} {Semantic relations between
  events and their time, locations and participants for event coreference
  resolution}.
\newblock In \emph{Proceedings of the International Conference Recent Advances
  in Natural Language Processing {RANLP} 2013}, pages 156--163, Hissar,
  Bulgaria. INCOMA Ltd. Shoumen, BULGARIA.

\bibitem[{Cybulska and Vossen(2014)}]{cybulska-vossen-2014-using}
Agata Cybulska and Piek Vossen. 2014.
\newblock \href
  {http://www.lrec-conf.org/proceedings/lrec2014/pdf/840_Paper.pdf} {Using a
  sledgehammer to crack a nut? lexical diversity and event coreference
  resolution}.
\newblock In \emph{Proceedings of the Ninth International Conference on
  Language Resources and Evaluation ({LREC}'14)}, pages 4545--4552, Reykjavik,
  Iceland. European Language Resources Association (ELRA).

\bibitem[{Davidson(1969)}]{Davidson1969}
Donald Davidson. 1969.
\newblock \href {https://doi.org/10.1007/978-94-017-1466-2_11} {\emph{The
  Individuation of Events}}, pages 216--234. Springer Netherlands, Dordrecht.

\bibitem[{Denis and Baldridge(2009)}]{515751742011}
Pascal Denis and Jason Baldridge. 2009.
\newblock \href {https://www.redalyc.org/articulo.oa?id=515751742011} {Global
  joint models for coreference resolution and named entity classification}.
\newblock \emph{Procesamiento del Lenguaje Natural}.

\bibitem[{Devlin et~al.(2019)Devlin, Chang, Lee, and
  Toutanova}]{devlin-etal-2019-bert}
Jacob Devlin, Ming-Wei Chang, Kenton Lee, and Kristina Toutanova. 2019.
\newblock \href {https://doi.org/10.18653/v1/N19-1423} {{BERT}: Pre-training of
  deep bidirectional transformers for language understanding}.
\newblock In \emph{Proceedings of the 2019 Conference of the North {A}merican
  Chapter of the Association for Computational Linguistics: Human Language
  Technologies, Volume 1 (Long and Short Papers)}, pages 4171--4186,
  Minneapolis, Minnesota. Association for Computational Linguistics.

\bibitem[{Flanigan et~al.(2014)Flanigan, Thomson, Carbonell, Dyer, and
  Smith}]{flanigan-etal-2014-discriminative}
Jeffrey Flanigan, Sam Thomson, Jaime Carbonell, Chris Dyer, and Noah~A. Smith.
  2014.
\newblock \href {https://doi.org/10.3115/v1/P14-1134} {A discriminative
  graph-based parser for the {A}bstract {M}eaning {R}epresentation}.
\newblock In \emph{Proceedings of the 52nd Annual Meeting of the Association
  for Computational Linguistics (Volume 1: Long Papers)}, pages 1426--1436,
  Baltimore, Maryland. Association for Computational Linguistics.

\bibitem[{Guarino et~al.(2022)Guarino, Baratella, and
  Guizzardi}]{Guarino2022-GUAETN}
Nicola Guarino, Riccardo Baratella, and Giancarlo Guizzardi. 2022.
\newblock \href {https://doi.org/10.3233/ao-220261} {Events, their names, and
  their synchronic structure}.
\newblock \emph{Applied ontology}, 17(2):249--283.

\bibitem[{Held et~al.(2021)Held, Iter, and Jurafsky}]{held-etal-2021-focus}
William Held, Dan Iter, and Dan Jurafsky. 2021.
\newblock \href {https://doi.org/10.18653/v1/2021.emnlp-main.106} {Focus on
  what matters: Applying discourse coherence theory to cross document
  coreference}.
\newblock In \emph{Proceedings of the 2021 Conference on Empirical Methods in
  Natural Language Processing}, pages 1406--1417, Online and Punta Cana,
  Dominican Republic. Association for Computational Linguistics.

\bibitem[{Kenyon-Dean et~al.(2018)Kenyon-Dean, Cheung, and
  Precup}]{kenyon2018resolving}
Kian Kenyon-Dean, Jackie Chi~Kit Cheung, and Doina Precup. 2018.
\newblock \href {https://arxiv.org/abs/1805.10985} {Resolving event coreference
  with supervised representation learning and clustering-oriented
  regularization}.
\newblock \emph{arXiv preprint arXiv:1805.10985}.

\bibitem[{Kojima et~al.(2023)Kojima, Gu, Reid, Matsuo, and
  Iwasawa}]{kojima2023large}
Takeshi Kojima, Shixiang~Shane Gu, Machel Reid, Yutaka Matsuo, and Yusuke
  Iwasawa. 2023.
\newblock \href {http://arxiv.org/abs/2205.11916} {Large language models are
  zero-shot reasoners}.

\bibitem[{Li et~al.(2021)Li, Ji, and Han}]{li-etal-2021-document}
Sha Li, Heng Ji, and Jiawei Han. 2021.
\newblock \href {https://doi.org/10.18653/v1/2021.naacl-main.69}
  {Document-level event argument extraction by conditional generation}.
\newblock In \emph{Proceedings of the 2021 Conference of the North American
  Chapter of the Association for Computational Linguistics: Human Language
  Technologies}, pages 894--908, Online. Association for Computational
  Linguistics.

\bibitem[{Liu et~al.(2019)Liu, Ott, Goyal, Du, Joshi, Chen, Levy, Lewis,
  Zettlemoyer, and Stoyanov}]{liu2019roberta}
Yinhan Liu, Myle Ott, Naman Goyal, Jingfei Du, Mandar Joshi, Danqi Chen, Omer
  Levy, Mike Lewis, Luke Zettlemoyer, and Veselin Stoyanov. 2019.
\newblock \href {http://arxiv.org/abs/1907.11692} {Roberta: A robustly
  optimized bert pretraining approach}.

\bibitem[{Lombard(2019)}]{lombard2019events}
Lawrence~Brian Lombard. 2019.
\newblock \href {https://doi.org/10.4324/9780429202179} {\emph{Events: A
  metaphysical study}}.
\newblock Routledge.

\bibitem[{Luo(2005)}]{10.3115/1220575.1220579}
Xiaoqiang Luo. 2005.
\newblock \href {https://doi.org/10.3115/1220575.1220579} {On coreference
  resolution performance metrics}.
\newblock In \emph{Proceedings of the Conference on Human Language Technology
  and Empirical Methods in Natural Language Processing}, HLT ’05, page
  25–32, USA. Association for Computational Linguistics.

\bibitem[{Luo et~al.(2014)Luo, Pradhan, Recasens, and
  Hovy}]{luo-etal-2014-extension}
Xiaoqiang Luo, Sameer Pradhan, Marta Recasens, and Eduard Hovy. 2014.
\newblock \href {https://doi.org/10.3115/v1/P14-2005} {An extension of {BLANC}
  to system mentions}.
\newblock In \emph{Proceedings of the 52nd Annual Meeting of the Association
  for Computational Linguistics (Volume 2: Short Papers)}, pages 24--29,
  Baltimore, Maryland. Association for Computational Linguistics.

\bibitem[{Magar and Schwartz(2022)}]{magar-schwartz-2022-data}
Inbal Magar and Roy Schwartz. 2022.
\newblock \href {https://doi.org/10.18653/v1/2022.acl-short.18} {Data
  contamination: From memorization to exploitation}.
\newblock In \emph{Proceedings of the 60th Annual Meeting of the Association
  for Computational Linguistics (Volume 2: Short Papers)}, pages 157--165,
  Dublin, Ireland. Association for Computational Linguistics.

\bibitem[{Moosavi et~al.(2019)Moosavi, Born, Poesio, and
  Strube}]{moosavi-etal-2019-using}
Nafise~Sadat Moosavi, Leo Born, Massimo Poesio, and Michael Strube. 2019.
\newblock \href {https://doi.org/10.18653/v1/P19-1408} {Using automatically
  extracted minimum spans to disentangle coreference evaluation from boundary
  detection}.
\newblock In \emph{Proceedings of the 57th Annual Meeting of the Association
  for Computational Linguistics}, pages 4168--4178, Florence, Italy.
  Association for Computational Linguistics.

\bibitem[{O{'}Gorman et~al.(2016)O{'}Gorman, Wright-Bettner, and
  Palmer}]{ogorman-etal-2016-richer}
Tim O{'}Gorman, Kristin Wright-Bettner, and Martha Palmer. 2016.
\newblock \href {https://doi.org/10.18653/v1/W16-5706} {Richer event
  description: Integrating event coreference with temporal, causal and bridging
  annotation}.
\newblock In \emph{Proceedings of the 2nd Workshop on Computing News Storylines
  ({CNS} 2016)}, pages 47--56, Austin, Texas. Association for Computational
  Linguistics.

\bibitem[{OpenAI(2023)}]{openai2023gpt4}
OpenAI. 2023.
\newblock \href {http://arxiv.org/abs/2303.08774} {Gpt-4 technical report}.

\bibitem[{Palmer et~al.(2005)Palmer, Gildea, and
  Kingsbury}]{palmer-etal-2005-proposition}
Martha Palmer, Daniel Gildea, and Paul Kingsbury. 2005.
\newblock \href {https://doi.org/10.1162/0891201053630264} {The {P}roposition
  {B}ank: An annotated corpus of semantic roles}.
\newblock \emph{Computational Linguistics}, 31(1):71--106.

\bibitem[{Pradhan et~al.(2022)Pradhan, Bonn, Myers, Conger, O{'}gorman, Gung,
  Wright-bettner, and Palmer}]{pradhan-etal-2022-propbank}
Sameer Pradhan, Julia Bonn, Skatje Myers, Kathryn Conger, Tim O{'}gorman, James
  Gung, Kristin Wright-bettner, and Martha Palmer. 2022.
\newblock \href {https://doi.org/10.18653/v1/2022.starsem-1.24} {{P}rop{B}ank
  comes of {A}ge{---}{L}arger, smarter, and more diverse}.
\newblock In \emph{Proceedings of the 11th Joint Conference on Lexical and
  Computational Semantics}, pages 278--288, Seattle, Washington. Association
  for Computational Linguistics.

\bibitem[{Pradhan et~al.(2014)Pradhan, Luo, Recasens, Hovy, Ng, and
  Strube}]{pradhan-EtAl:2014:P14-2}
Sameer Pradhan, Xiaoqiang Luo, Marta Recasens, Eduard Hovy, Vincent Ng, and
  Michael Strube. 2014.
\newblock \href {http://www.aclweb.org/anthology/P14-2006} {Scoring coreference
  partitions of predicted mentions: A reference implementation}.
\newblock In \emph{Proceedings of the 52nd Annual Meeting of the Association
  for Computational Linguistics (Volume 2: Short Papers)}, pages 30--35,
  Baltimore, Maryland. Association for Computational Linguistics.

\bibitem[{Radford et~al.(2018)Radford, Narasimhan, Salimans, Sutskever
  et~al.}]{radford2018improving}
Alec Radford, Karthik Narasimhan, Tim Salimans, Ilya Sutskever, et~al. 2018.
\newblock \href {https://www.mikecaptain.com/resources/pdf/GPT-1.pdf}
  {Improving language understanding by generative pre-training}.

\bibitem[{Radford et~al.(2019)Radford, Wu, Child, Luan, Amodei, Sutskever
  et~al.}]{radford2019language}
Alec Radford, Jeffrey Wu, Rewon Child, David Luan, Dario Amodei, Ilya
  Sutskever, et~al. 2019.
\newblock \href
  {https://blocksml.com/genaipapers/Language%20Models%20are%20Unsupervised%20Multitask%20Learners.pdf}
  {Language models are unsupervised multitask learners}.
\newblock \emph{OpenAI blog}, 1(8):9.

\bibitem[{Schuler(2005)}]{schuler2005verbnet}
Karin~Kipper Schuler. 2005.
\newblock \href {https://doi.org/10.5555/1104493} {\emph{VerbNet: A
  broad-coverage, comprehensive verb lexicon}}.
\newblock University of Pennsylvania.

\bibitem[{Song et~al.(2018)Song, Bies, Mott, Li, Strassel, and
  Caruso}]{song-etal-2018-cross}
Zhiyi Song, Ann Bies, Justin Mott, Xuansong Li, Stephanie Strassel, and
  Christopher Caruso. 2018.
\newblock \href {https://aclanthology.org/L18-1558} {Cross-document,
  cross-language event coreference annotation using event hoppers}.
\newblock In \emph{Proceedings of the Eleventh International Conference on
  Language Resources and Evaluation ({LREC} 2018)}, Miyazaki, Japan. European
  Language Resources Association (ELRA).

\bibitem[{Song et~al.(2015)Song, Bies, Strassel, Riese, Mott, Ellis, Wright,
  Kulick, Ryant, and Ma}]{song-etal-2015-light}
Zhiyi Song, Ann Bies, Stephanie Strassel, Tom Riese, Justin Mott, Joe Ellis,
  Jonathan Wright, Seth Kulick, Neville Ryant, and Xiaoyi Ma. 2015.
\newblock \href {https://doi.org/10.3115/v1/W15-0812} {From light to rich
  {ERE}: Annotation of entities, relations, and events}.
\newblock In \emph{Proceedings of the The 3rd Workshop on {EVENTS}: Definition,
  Detection, Coreference, and Representation}, pages 89--98, Denver, Colorado.
  Association for Computational Linguistics.

\bibitem[{Spaulding et~al.(2024)Spaulding, Conger, Gershman, Morshed, Brown,
  Pustejovsky, Uceda-Sosa, Ge, and Palmer}]{spaulding-etal-2024-propbank}
Elizabeth Spaulding, Kathryn Conger, Anatole Gershman, Mahir Morshed,
  Susan~Windisch Brown, James Pustejovsky, Rosario Uceda-Sosa, Sijia Ge, and
  Martha Palmer. 2024.
\newblock \href {https://aclanthology.org/2024.law-1.16} {{P}rop{B}ank goes
  public: Incorporation into {W}ikidata}.
\newblock In \emph{Proceedings of The 18th Linguistic Annotation Workshop
  (LAW-XVIII)}, pages 166--175, St. Julians, Malta. Association for
  Computational Linguistics.

\bibitem[{Touvron et~al.(2023)Touvron, Martin, Stone, Albert, Almahairi,
  Babaei, Bashlykov, Batra, Bhargava, Bhosale et~al.}]{touvron2023llama}
Hugo Touvron, Louis Martin, Kevin Stone, Peter Albert, Amjad Almahairi, Yasmine
  Babaei, Nikolay Bashlykov, Soumya Batra, Prajjwal Bhargava, Shruti Bhosale,
  et~al. 2023.
\newblock \href {https://arxiv.org/abs/2307.09288} {Llama 2: Open foundation
  and fine-tuned chat models}.
\newblock \emph{arXiv preprint arXiv:2307.09288}.

\bibitem[{Vaswani et~al.(2017)Vaswani, Shazeer, Parmar, Uszkoreit, Jones,
  Gomez, Kaiser, and Polosukhin}]{NIPS2017_3f5ee243}
Ashish Vaswani, Noam Shazeer, Niki Parmar, Jakob Uszkoreit, Llion Jones,
  Aidan~N Gomez, \L~ukasz Kaiser, and Illia Polosukhin. 2017.
\newblock \href
  {https://proceedings.neurips.cc/paper_files/paper/2017/file/3f5ee243547dee91fbd053c1c4a845aa-Paper.pdf}
  {Attention is all you need}.
\newblock In \emph{Advances in Neural Information Processing Systems},
  volume~30. Curran Associates, Inc.

\bibitem[{Vilain et~al.(1995)Vilain, Burger, Aberdeen, Connolly, and
  Hirschman}]{10.3115/1072399.1072405}
Marc Vilain, John Burger, John Aberdeen, Dennis Connolly, and Lynette
  Hirschman. 1995.
\newblock \href {https://doi.org/10.3115/1072399.1072405} {A model-theoretic
  coreference scoring scheme}.
\newblock In \emph{Proceedings of the 6th Conference on Message Understanding},
  MUC6 ’95, page 45–52, USA. Association for Computational Linguistics.

\bibitem[{Wang et~al.(2020)Wang, Wang, Han, Jiang, Han, Liu, Li, Li, Lin, and
  Zhou}]{wang-etal-2020-maven}
Xiaozhi Wang, Ziqi Wang, Xu~Han, Wangyi Jiang, Rong Han, Zhiyuan Liu, Juanzi
  Li, Peng Li, Yankai Lin, and Jie Zhou. 2020.
\newblock \href {https://doi.org/10.18653/v1/2020.emnlp-main.129} {{MAVEN}: {A}
  {M}assive {G}eneral {D}omain {E}vent {D}etection {D}ataset}.
\newblock In \emph{Proceedings of the 2020 Conference on Empirical Methods in
  Natural Language Processing (EMNLP)}, pages 1652--1671, Online. Association
  for Computational Linguistics.

\bibitem[{Wei et~al.(2022)Wei, Wang, Schuurmans, Bosma, ichter, Xia, Chi, Le,
  and Zhou}]{NEURIPS2022_9d560961}
Jason Wei, Xuezhi Wang, Dale Schuurmans, Maarten Bosma, brian ichter, Fei Xia,
  Ed~Chi, Quoc~V Le, and Denny Zhou. 2022.
\newblock \href
  {https://proceedings.neurips.cc/paper_files/paper/2022/file/9d5609613524ecf4f15af0f7b31abca4-Paper-Conference.pdf}
  {Chain-of-thought prompting elicits reasoning in large language models}.
\newblock In \emph{Advances in Neural Information Processing Systems},
  volume~35, pages 24824--24837. Curran Associates, Inc.

\bibitem[{Wright-Bettner et~al.(2019)Wright-Bettner, Palmer, Savova, de~Groen,
  and Miller}]{wright-bettner-etal-2019-cross}
Kristin Wright-Bettner, Martha Palmer, Guergana Savova, Piet de~Groen, and
  Timothy Miller. 2019.
\newblock \href {https://doi.org/10.18653/v1/D19-6201} {Cross-document
  coreference: An approach to capturing coreference without context}.
\newblock In \emph{Proceedings of the Tenth International Workshop on Health
  Text Mining and Information Analysis (LOUHI 2019)}, pages 1--10, Hong Kong.
  Association for Computational Linguistics.

\bibitem[{Wu et~al.(2023)Wu, Qiu, Ross, Aky{\"u}rek, Chen, Wang, Kim, Andreas,
  and Kim}]{wu2023reasoning}
Zhaofeng Wu, Linlu Qiu, Alexis Ross, Ekin Aky{\"u}rek, Boyuan Chen, Bailin
  Wang, Najoung Kim, Jacob Andreas, and Yoon Kim. 2023.
\newblock \href {https://arxiv.org/abs/2307.02477} {Reasoning or reciting?
  exploring the capabilities and limitations of language models through
  counterfactual tasks}.
\newblock \emph{arXiv preprint arXiv:2307.02477}.

\bibitem[{Xu et~al.(2022)Xu, Wang, Liu, Zeng, Chang, and
  Sui}]{xu-etal-2022-two}
Runxin Xu, Peiyi Wang, Tianyu Liu, Shuang Zeng, Baobao Chang, and Zhifang Sui.
  2022.
\newblock \href {https://doi.org/10.18653/v1/2022.naacl-main.370} {A two-stream
  {AMR}-enhanced model for document-level event argument extraction}.
\newblock In \emph{Proceedings of the 2022 Conference of the North American
  Chapter of the Association for Computational Linguistics: Human Language
  Technologies}, pages 5025--5036, Seattle, United States. Association for
  Computational Linguistics.

\bibitem[{Yang et~al.(2022{\natexlab{a}})Yang, Lu, and
  Petzold}]{Yang2022FewShotDE}
Xianjun Yang, Yujie Lu, and Linda Petzold. 2022{\natexlab{a}}.
\newblock \href {https://arxiv.org/abs/2209.02203} {Few-shot document-level
  event argument extraction}.
\newblock \emph{ArXiv}, abs/2209.02203.

\bibitem[{Yang et~al.(2022{\natexlab{b}})Yang, Peynetti, Meerman, and
  Tanner}]{yang-etal-2022-gpt}
Xiaohan Yang, Eduardo Peynetti, Vasco Meerman, and Chris Tanner.
  2022{\natexlab{b}}.
\newblock \href {https://doi.org/10.18653/v1/2022.insights-1.10} {What {GPT}
  knows about who is who}.
\newblock In \emph{Proceedings of the Third Workshop on Insights from Negative
  Results in NLP}, pages 75--81, Dublin, Ireland. Association for Computational
  Linguistics.

\bibitem[{Zhang et~al.(2022)Zhang, Strubell, and
  Hovy}]{zhang-etal-2022-transfer}
Zhisong Zhang, Emma Strubell, and Eduard Hovy. 2022.
\newblock \href {https://aclanthology.org/2022.emnlp-main.169} {Transfer
  learning from semantic role labeling to event argument extraction with
  template-based slot querying}.
\newblock In \emph{Proceedings of the 2022 Conference on Empirical Methods in
  Natural Language Processing}, pages 2627--2647, Abu Dhabi, United Arab
  Emirates. Association for Computational Linguistics.

\end{thebibliography}
\bibliographystyle{lrec-coling2024-natbib}

\end{document}